\documentclass[11pt,a4paper]{article}
\usepackage[hyperref]{acl2020}
\usepackage{graphicx} 
\usepackage{times}
\usepackage{latexsym}

\usepackage{microtype}

\aclfinalcopy 

\title{Modeling Discourse Structure for Document-level Neural Machine Translation}

\author{Junxuan Chen$^{1}\thanks{\ \ This work is done when Junxuan Chen was interning at Xiaomi AI Lab, Xiaomi Inc., Beijing, China.}$, \ Xiang Li$^{2}$, \  Jiarui Zhang$^{1}$, \
Chulun Zhou$^{1}$, \\ {\bf Jianwei Cui}$^{2}$, \ {\bf Bin Wang}$^{2}$, \ {\bf Jinsong Su}$^{1}\thanks{\ \ Corresponding author.}$ \\
$^{1}$Xiamen University, Xiamen, China \\ $^{2}$Xiaomi AI Lab, Xiaomi Inc., Beijing, China \\
 {\tt \{chenjx,zhangjiarui,clzhou\}@stu.xmu.edu.cn  jssu@xmu.edu.cn} \\ {\tt \{lixiang21,cuijianwei,wangbin11\}@xiaomi.com } \\ 
}
\date{}

\begin{document}
\maketitle
\begin{abstract}
Recently, document-level neural machine translation (NMT) has become a hot topic in the community of machine translation.
Despite its success, most of existing studies ignored the discourse structure information of the input document to be translated, which has shown effective in other tasks.
In this paper, we propose to improve document-level NMT with the aid of discourse structure information.
Our encoder is based on a hierarchical attention network (HAN) \cite{miculicich-etal-2018-document}.
Specifically, we first parse the input document to obtain its discourse structure.
Then, we introduce a Transformer-based path encoder to embed the discourse structure information of each word.
Finally, we combine the discourse structure information with the word embedding before it is fed into the encoder. 
Experimental results on the English-to-German dataset show that our model can significantly outperform both Transformer and Transformer+HAN.
\end{abstract}

\section{Introduction}
Neural machine translation (NMT) has made great progress in the past decade. 
In practical applications, the need for NMT systems has expanded from individual sentences to complete documents.
Therefore, document-level NMT has gradually drawn much more attention.
Contextual information is particularly important for obtaining high-quality document translation.
To get better contextual information, researchers have proposed many methods (e.g., memory network and hierarchical attention network) for document-level translation \cite{sim-smith-2017-integrating,tiedemann-scherrer-2017-neural,wang-etal-2017-exploiting-cross,tu-etal-2017-context,wang-etal-2017-exploiting-cross,voita-etal-2018-context,zhang-etal-2018-improving,miculicich-etal-2018-document,maruf-haffari-2018-document,maruf-etal-2019-selective,yang-etal-2019-enhancing-context}.
Discourse structure, as well as raw contextual sentences, is a major component of the document.
And it has been proved to be effective in many other tasks, such as automatic document summarization \cite{yoshida-etal-2014-dependency,isonuma-etal-2019-unsupervised} and sentiment classification \cite{schouten-frasincar-2016-commit,nejat-etal-2017-exploring}. 
However, to the best of our knowledge, discourse structure has not been explicitly used in document-level NMT.

To address the above problem, we propose to improve document-level NMT with the aid of discourse structure information.
First, we represent each input document with a Rhetorical Structure Theory-based discourse tree \cite{Mann1988Rhetorical}. 
Then, we use a Transformer-based path encoder to embed the discourse structure path of each word and combine it with the corresponding word embedding before feeding it into the sentence encoder.
In this way, discourse structure information can be fully exploited to enrich word representations and guide the context encoder to capture the relevant context of the current sentence.
Finally, we adopt HAN \cite{miculicich-etal-2018-document} as our context encoder to model context information in a hierarchical manner.

Our contributions are as follows: 
(i) We propose a novel and efficient approach to explicitly exploit discourse structure information for document-level NMT. 
Particularly, our approach is applicable for any other context encoder of document-level NMT;
(ii) We carry out experiments on English-to-German translation task and experimental results show that our model outperforms competitive baselines.

\begin{figure*}
\centering
\includegraphics[width=15cm,height=8cm]{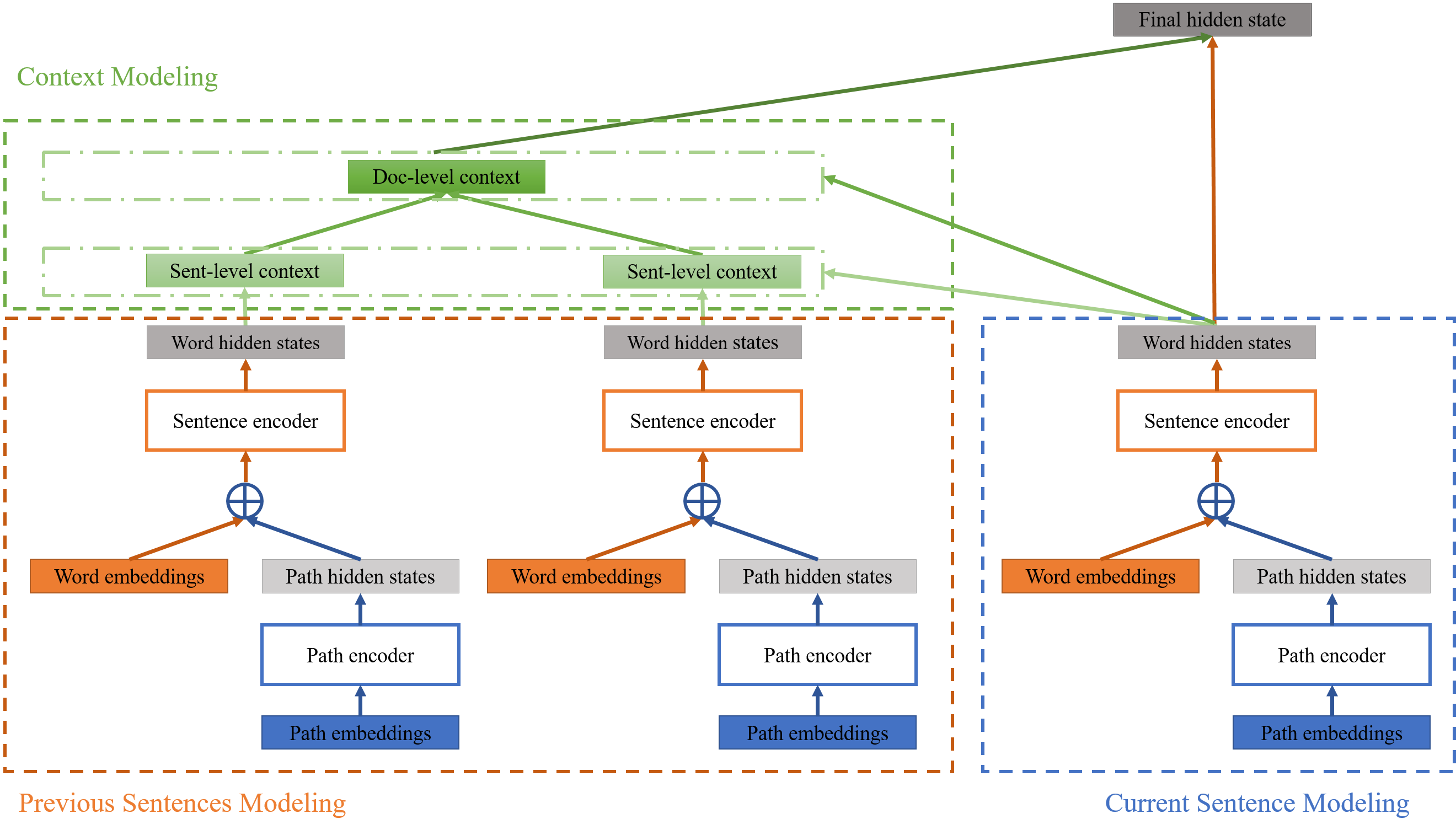}
\caption{The architecture of our proposed encoder}
\label{overarchitecture}
\end{figure*}

\section{Related Work}
In the era of statistical machine translation, 
document-level machine translation has become one of the research focuses in the community of machine translation. \cite{xiao2011document,su-etal-2012-translation,xiao-etal-2012-topic,su-etal-2015-context}.
Recently,
with the rapid development of NMT,
document-level NMT has also gradually attracted people's attention \cite{voita-etal-2018-context,maruf-haffari-2018-document,miculicich-etal-2018-document,maruf-etal-2019-selective,yang-etal-2019-enhancing-context}.
Typically, existing studies aim to improve document-level translation quality with the help of document context, which is usually extracted from neighboring sentences of the current sentence.
For example,some researchers applied cache-based models to selectively remember the most relevant context information of the document \cite{voita-etal-2018-context,maruf-haffari-2018-document,kuang-etal-2018-modeling}, while some researchers employed hierarchical context networks to catch document context information for Transformer \cite{miculicich-etal-2018-document,maruf-etal-2019-selective,yang-etal-2019-enhancing-context}.
Specifically,
Miculicich et al. \shortcite{miculicich-etal-2018-document} proposed a hierarchical attention network to model contextual information, Maruf et al. \shortcite{maruf-etal-2019-selective} applied a selective attention method to select contextual information that is more relevant to the current sentence, 
and Yang et al. \shortcite{yang-etal-2019-enhancing-context} employed capsule network to model multi-angle context information.

Although these methods have made some progress in document-level NMT, they all ignored the discourse structure information, which can be used to not only enrich word embedding but also guide the selection of relevant context for the current sentence.

\section{Our Encoder}
We propose a novel document-level NMT model based on HAN \cite{miculicich-etal-2018-document}.
The difference between ours and HAN lies in that we introduce the RST-based discourse structure to represent the document-level context, which is incorporated into HAN to refine translation.

Figure \ref{overarchitecture} gives the architecture of our proposed encoder.
In addition to the standard encoder for the current sentence, 
it contains HAN \cite{miculicich-etal-2018-document} as context encoder, and a novel path encoder for the discourse structure.
We first use the Transformer-based path encoder to model discourse structure information.
Then, we combine the embedding of each input word with its corresponding path embedding vector and feed the combined vector into the sentence encoder.
Finally, we use the hierarchical attention network to capture the global contextual embedding and update the hidden states of current sentence as the final output of the whole encoder.

\begin{figure*}
\centering
\includegraphics[width=15cm,height=5.7cm]{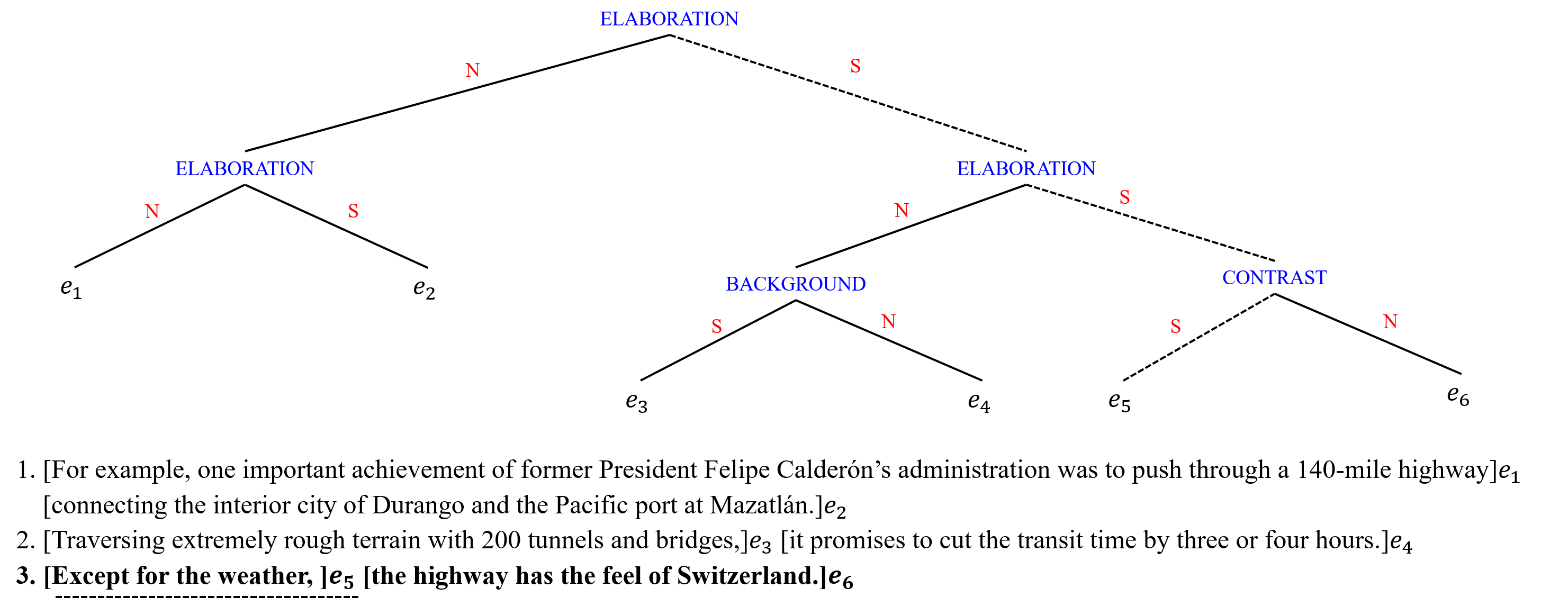}
\caption{An example discourse tree of with six EDUs. $N$ and $S$ denote the relative importance label \emph{NUCLEUS} and \emph{SATELLITE}, respectively. Sentence 3 is the current sentence to be translated, with two previous context sentences 1 and 2. On the tree, the path marked with dotted lines from the root node to the leaf node $e_5$ is used to represent the discourse structure of $e_5$.}
\label{RST}
\end{figure*}

In our model, the translation of a document is made by translating each of its sentences sequentially.
We introduce discourse structure for both the current sentence and contextual sentences.
Given a source document $X$, the translation probability of the target document $Y$ can be defined as:
\begin{equation}P(Y \mid X;\theta)=\prod_{j=1}^J P(Y^j \mid X^j,D^j,S;\theta)\label{docnmtDS},\end{equation}
where $X^j$ and $Y^j$ denote the $j$-th source sentence and its target translation respectively, $D^j$ denotes the contextual sentences, $S$ represents the discourse structure of the document to be translated, and $\theta$ is the parameter set of the model. 

\subsection{RST-based Discourse Structure}
For each document to be translated,
we parse it to obtain its discourse structure based on Rhetorical Structure Theory (RST) \cite{Mann1988Rhetorical}.
RST is one of the most influential theories of document coherence.
According to RST, we represent each document with a hierarchical tree.
As shown in Figure \ref{RST}, the discourse tree contains some leaf nodes, each of which indicates an Elementary Discourse Unit (EDU). 
The adjacent leaf nodes are recursively connected into units by certain coherence relations (e.g., \emph{ELABORATION, BACKGROUND}) until the entire tree is built. 
Besides, \emph{NUCLEUS} or \emph{SATELLITE} is used to mark the relative importance of child node units in the relationship.

In this work, we represent the discourse structure information of each word using its discourse path from root node to its corresponding leaf node.
Each path is a mixed label sequence composed of the discourse relationship and the importance label (e.g., \emph{NUCLEUS\_ELABORATION, SATELLITE\_BACKGROUND}).
Please note that all tokens in the same EDU share the same discourse structure.
For example, the discourse structure of EDU $e_5$ is "\emph{SATELLITE\_ELABORATION SATELLITE\_ELABORATION SATELLITE\_CONTRAST}".

\subsection{Discourse Structure Path Embedding}
To integrate the structural information of words into the our HAN-based document-level NMT model,
we first additionaly introduce a Transformer-based path encoder to encode discourse structure paths of words.
Specifically, for each word $w_i$, we directly consider its discourse structure path $p_i$ as a sequence and then employ the path encoder to learn its contextual hidden states,
which can be finally averaged to produce the overall discourse embedding vector $d_i$.
Then, we enrich each input word embedding with its corresponding discourse embedding vector before it is fed into the context encoder or the translation encoder.
Concretely, for the word $w_i$, we define its enriched vector as the sum of its word embedding and discourse embedding:
$\widetilde{x_i}=x_i+d_i$.

\subsection{HAN-based Context Modeling}
Following \shortcite{miculicich-etal-2018-document}, 
we apply hierarchical attention network (HAN) as our context encoder. 
Due to the advantage of accurately capturing different levels of contexts,
HAN has been widely used in many tasks, such as document classification \cite{yang-etal-2016-hierarchical}, stance detection \cite{sun-etal-2018-stance}, sentence-level NMT \cite{su2018hierarchy}.
Using this encoder,
we mainly focus on two levels of context modeling:

\paragraph{Sentence-level Context Modeling}
For the $i$-th word of the current sentence, 
we employ muti-head attention \cite{DBLP:journals/corr/VaswaniSPUJGKP17} to summarize the context from the $k$-th context sentence:
\begin{equation}
{cs}_{i,k} = {\rm MultiHead}(f_s(h_i), \mathbf{H_k}), \end{equation}
where $f_s$ is a linear transformation function, $h_i$ denotes the hidden state representation of the $i$-th token of current sentence.
By doing so, our context encoder can exploit different types of relation between words to better capture sentence-level context.
And $\mathbf{H_k}$ is the hidden state representation of the $k$-th context sentence and is used as value and key for attention. 

\paragraph{Document-level Context Modeling}
Unlike the above modeling, here we mainly on capturing the context information from previous $K$ sentences for the $i$-th word of the current sentence.
\begin{equation}{cd}_i = {\rm FFN}({\rm MultiHead}(f_d(h_i), \mathbf{CS_i})), \end{equation}
where $f_d$ is a linear transformation, and $\mathbf{CS_i} = [cs_{i, 1}, cs_{i, 2}, {\cdots}, cs_{i, K}]$ is the sentence-level context of $K$ contextual sentences.

\paragraph{Integrating Document-level Context into the Translation Encoder}
Finally, we integrate the above-mentioned document-level context into the translation encoder via a gating operation:
\begin{equation}{\lambda}_i = {\sigma}(\mathbf{W_h} h_i + \mathbf{W_{cd}} {cd}_i)\end{equation}
\begin{equation}\widetilde{h_i} = {\lambda}_i h_i + (1-{\lambda}_i) {cd}_i \end{equation}
where $\mathbf{W_h}$ and $\mathbf{W_{cd}}$ denote parameter matrices for $h_i$ and $cd_i$, and $\widetilde{h_i}$ is the final output of the encoder.

\section{Experiments}
\subsection{Settings}
\paragraph{Datasets}
We conduct our experiments on English-to-German translation task. 
The sentence-aligned document-delimited News Comment v11 corpus \footnote{http://www.casmacat.eu/corpus/news-commentary.html}, the WMT16 newstest2015 and the newstest2016 are used as the training set, development and test set, respectively.

We download all the above corpus from \cite{maruf-etal-2019-selective}, of which statistics are provided in Table \ref{dataset}.

\begin{table}
\centering
\begin{tabular}{lrr}
\hline & \textbf{\#Sentences} & \textbf{Document Len} \\ \hline
Training & 236,287 & 38.93 \\
Development & 2,169 & 26.78 \\
Test & 2,999 & 19.35 \\
\hline
\end{tabular}
\caption{The statistical of our datasets.
\#Sentence indicates the number of sentences, and Document length means the average number of sentences in document.}
\label{dataset} 
\end{table}

\paragraph{Settings}
We use Transformer \cite{DBLP:journals/corr/VaswaniSPUJGKP17} as our context-agnostic baseline system and Transformer+HAN \cite{miculicich-etal-2018-document} as our context-aware baseline system.
We conduct experiments using the same configuration as HAN.
Specifically, both sentence encoder and decoder are composed of 6 hidden layers, while path encoder is composed of 2 hidden layers.
We use three previous sentences as contextual sentences for current sentence.
The hidden size and point-wise FFN size are 512 and 2048 respectively. 
The dropout rates for all hidden states are set to 0.1.
The source and target vocabulary sizes are both 30K.
At the training phase, we use the Adam optimizer \cite{Kingma2015AdamAM} and the batch sizes of context-agnostic model and context-aware model are 4096 and 1024, respectively.
Finally, we use case-sensitive BLEU \cite{papineni-etal-2002-bleu} and TER \cite{snover2006study} to measure the translation quality.

\paragraph{Data Preprocessing}
All datasets are tokenized and truecased 
using the scripts of Moses Toolkit \cite{koehn-etal-2007-moses}.
We split them into subword units using a joint bye pair encoding model with 30K merge operations. 
To get discourse structure of the input documents, we first apply the open-source tool NeuralEDUSeg \cite{wang-etal-2018-toward} obtaining non-overlapping EDUs.
Then, we employ StageDP \cite{wang-etal-2017-two} to obtain discourse structure trees of segmented documents.
Afterwards, we extract the path from root node to leaf node as the discourse structure information for the corresponding EDU, where all words share the same discourse structure path.

\begin{table*}
\centering
\setlength{\tabcolsep}{11mm}{
\begin{tabular}{lll}
\hline \textbf{Model} & \textbf{BLEU} & \textbf{TER} \\ \hline
Transformer & 22.78 & 59.3 \\
Transformer+DS & 23.61 (+0.83) & 58.5 (-0.8) \\
Transformer+HAN & 24.45 (+1.67) & 56.9 (-2.4) \\
Transformer+HAN+DS & \textbf{24.84} (+2.06) & \textbf{56.4} (-2.9) \\
\hline
\end{tabular}}
\caption{ BLEU and TER scores for different models. The best scores are marked in bold. HAN denotes Hierarchical Attention Network which is used to get context information while DS denotes Discourse Structure information.}
\label{mainresult}
\end{table*}

\subsection{Results and Analysis}
Table \ref{mainresult} shows the experimental results for different models. 
The sentence-level Transformer, context-agnostic baseline, obtains a result of 22.78 BLEU and 59.3 TER, while the context-aware baseline Transformer+HAN \cite{miculicich-etal-2018-document} obtains 24.45 BLEU and 56.9 TER.
The sentence-level Transformer integrated with discourse structure achieves an improvement of 0.83 on BLEU and a decline of 0.8 on TER.
By contrast, our model integrated with contextual information and discourse structure information further gains a better performance, 2.06 higher than Transformer and 0.39 higher than Transformer+HAN on BLEU, 2.9 lower than Transformer and 0.5 lower than Transformer+HAN on TER.

Our experimental results show that discourse structure features indeed provide helpful information to enhance the translation quality of both context-agnostic and context-aware document-level NMT models.
Please note that our approach is also applicable to other document-level NMT models.

\section{Conclusion}
This paper has presented a novel discourse structure-based encoder for document-level NMT. 
The main idea of our encoder lies in enriching the input word embeddings with their path embeddings based on discourse structure.
Experimental results on English-to-German translation verify the effectiveness of our proposed encoder.

In the future, we plan to extend our encoder to other NLP tasks, such as simultaneous translation.
Simultaneous translation, as well as document-level NMT, has difficulty in modeling the long context so that it may be effective to improve the re-translation with the structure information modeled by our encoder.
Finally, we will focus on refining document-level NMT with variational neural networks, which has shown effecitive in previous studies of sentence-level NMT \cite{zhang-etal-2016-variational-neural,su2018variational}.

\section*{Acknowledgments}
This work was supported by the National Key R\&D Program of China under Grant 2019QY1803, the National Natural Science Foundation of China (No. 61672440), and the Scientific Research Project of National Language Committee of China (No. YB135-49).

\bibliography{anthology,acl2020}
\bibliographystyle{acl_natbib}

\end{document}